\documentclass[10pt,twocolumn,letterpaper]{article}

\usepackage{cvpr}              

\usepackage[dvipsnames]{xcolor}

\definecolor{cvprblue}{rgb}{0.21,0.49,0.74}
\usepackage[pagebackref,breaklinks,colorlinks,citecolor=cvprblue]{hyperref}
\usepackage{multirow}
\usepackage[accsupp]{axessibility}

\def\paperID{1} 
\def\confName{CVPR}
\def\confYear{2024}

\title{Unified Physical-Digital Attack Detection Challenge}
\newcommand*\samethanks[1][\value{footnote}]{\footnotemark[#1]}

\author{
Haocheng Yuan$^{\rm 1}$\thanks{Equal contribution.}, 
Ajian Liu$^{\rm 2}$\samethanks, 
Junze Zheng$^{\rm 1}$, 
Jun Wan$^{\rm 1, 2, 6}$\thanks{Corresponding author.}\\
Jiankang Deng$^{\rm 3}$, 
Sergio Escalera$^{\rm 4}$,
Hugo Jair Escalante$^{\rm 7, 8}$, 
Isabelle Guyon$^{\rm 8}$, 
Zhen Lei$^{\rm 2,5,6}$ \\
$^{\rm 1}$M.U.S.T, Macau; 
$^{\rm 2}$MAIS, CASIA, China; 
$^{\rm 3}$ICL, UK; 
$^{\rm 4}$CVC, Spain; 
$^{\rm 5}$CAIR, HKISI, CAS; \\
$^{\rm 6}$SAI, UCAS, China; 
$^{\rm 7}$INAOE, CINVESTAV, Mexico;
$^{\rm 8}$ChaLearn, USA \\
\tt\footnotesize
$^1$Allen.hcyuan@outlook.com, 
\tt\footnotesize
$^2$\{ajian.liu,jun.wan\}@ia.ac.cn, 
\tt\footnotesize
$^2$zlei@nlpr.ia.ac.cn 
}

\begin{document}
\maketitle
\begin{abstract}
Face Anti-Spoofing (FAS) is crucial to safeguard Face Recognition (FR) Systems. In real-world scenarios, FRs are confronted with both physical and digital attacks. However, existing algorithms often address only one type of attack at a time, which poses significant limitations in real-world scenarios where FR systems face hybrid physical-digital threats. To facilitate the research of Unified Attack Detection (UAD) algorithms, a large-scale UniAttackData dataset has been collected. UniAttackData is the largest public dataset for Unified Attack Detection, with a total of 28,706 videos, where each unique identity encompasses all advanced attack types. Based on this dataset, we organized a Unified Physical-Digital Face Attack Detection Challenge to boost the research in Unified Attack Detections. It attracted 136 teams for the development phase, with 13 qualifying for the final round. The results re-verified by the organizing team were used for the final ranking. This paper comprehensively reviews the challenge, detailing the dataset introduction, protocol definition, evaluation criteria, and a summary of published results. Finally, we focus on the detailed analysis of the highest-performing algorithms and offer potential directions for unified physical-digital attack detection inspired by this competition. Challenge Website:~\url{https://sites.google.com/view/face-anti-spoofing-challenge/welcome/challengecvpr2024}
\end{abstract}    
\section{Introduction}
\label{sec:intro}
\begin{table}[]
\scalebox{0.79}{
\begin{tabular}{|c|c|c|}
\hline
Ranking & Team Name         & Leader Name,   Affiliation                                                                                                                              \\ \hline
1       & MTFace            & Xianhua He, Meituan                                                                                                                                     \\ \hline
2       & SeaRecluse        & Minzhe Huang, Akuvox                                                                                                                                   \\ \hline
3       & duileduile        & \begin{tabular}[c]{@{}c@{}}Jiaruo   Yu, \\      INTSIG Information Co. Ltd\end{tabular}                                                                   \\ \hline
4       & BSP-Idiap         & \begin{tabular}[c]{@{}c@{}}Anjith   George,\\       Idiap Research Institute\end{tabular}                                                               \\ \hline
5       & VAI-Face          & Vu Minh Quan, Viettel   AI                                                                                                                              \\ \hline
6       & L\&L\&W           & \begin{tabular}[c]{@{}c@{}}Tongming   Wan,\\       Central South University\end{tabular}                                                                \\ \hline
7       & SARM              & Jun Lan, SARM                                                                                                                                           \\ \hline
8       & M2-Purdue         & Shu Hu, Purdue   University                                                                                                                             \\ \hline
9       & Cloud Recesses   & \begin{tabular}[c]{@{}c@{}}Peipeng   Yu, \\      Nanyang Technological Univeristy\end{tabular}                                                          \\ \hline
10      & ImageLab          & \begin{tabular}[c]{@{}c@{}}Sabari   Nathan, Couger Inc., \\      Sethu Institute of Technology, \\      Thiagarajar college of engineering\end{tabular} \\ \hline
11      & BOVIFOCR-UFPR     & \begin{tabular}[c]{@{}c@{}}Bernardo   Biesseck, \\      Federal University of Paraná\end{tabular}                                                       \\ \hline
12      & Inria-CENATAV-Tec & \begin{tabular}[c]{@{}c@{}}Luis   Santiago Luévano García, \\      Inria, CENATAV, \\      Tecnologico de Monterrey\end{tabular}                        \\ \hline
13      & Vicognit          & Manoj Sharma, Bennett   Univeristy                                                                                                                      \\ \hline
\end{tabular}
}
\caption{List of the team and affiliation names in the final ranking
of this challenge.}
\label{Tab1_ranks}
\end{table}

Face Anti-Spoofing (FAS) is essential to ensure the security of Face Recognition (FR) systems by identifying whether the image is live or fake. Attacks and corresponding detection methods can be classified into physical and digital categories. Physical Attacks (PA) involve the presentation of face replicas, like prints, masks, and screen replies. With the release of several high-quality 2D datasets~\cite{2Ddataset1_diverse, 2Ddataset3_SiW, zhang2019dataset}, and 3D-Mask datasets~\cite{3Ddataset2_3DMAD-SUP, 3Ddataset3_Rose-Youtu, 3Ddataset4_SMAD, 3Ddataset5_WMCA}, existing works~\cite{PAD2_Distangle, PAD4_AdapNor, PAD6_3dpcnet, PAD7_MAViT, PAD8_AdvCrsModa, PAD9_Distngle, liu2022spoof} demonstrated a satisfying result, while other efforts~\cite{MuModal2_centralDif, MuModal4_FMViT, MuModal5_SDfusion} have concentrated on leveraging multi-modal information to uncover spoofing clues. Wang et al.~\cite{wang2024multi} introduce a Multi-Domain Incremental Learning (MDIL) approach for Presentation Attack Detection (PAD), effectively acquiring new domain knowledge while preserving performance across previously learned domains. Digital Attacks (DA) aim to manipulate faces before they are presented for verification, involving Deepfakes~\cite{Digital1_facedancer, Digital2_InsightFace, Digital4_safa, Digital5_OneShotTH, Digital6_DaGAN} and adversarial attacks~\cite{Digitaladv4_fgtm, Digitaladv6_ssah}. With the release of the dataset~\cite{Digtalds1_faceswap, Digitalds2_face2face, Digitalds3_neuraltex}, promising results have been obtained by current works~\cite{DAD1_ffpp, DAD2_mulattention, DAD3_frequencyA, DAD4_wilddf, DAD5_icTrans, DAD6_AADfas, liu2019deep, huang2022adaptive} through distinguishing digitally manipulated facial artifacts. Recent work~\cite{liu2024cfpl} utilizes large-scale VLMs and text features to dynamically adjust classifier weights, enhancing the exploration of generalizable visual features.

However, related studies continue to investigate PA Detection (PAD) and DA Detection (DAD) as distinct tasks, leading to extensive computing resources consumed during deployment. We identify two main reasons for the scarcities of Unified Attack Detection (UAD): (1) Lack of a large-scale dataset that unifies both physical and digital attacks. While two notable datasets, GrandFake~\cite{UniDataset1_Grandfake} and JFSFDB~\cite{UniDataset2_JFSFDB} have been introduced to address the UAD challenge, their approach primarily combines PA and DA datasets. (2) Lack of a public baseline benchmark measuring the UAD algorithms. Although physical and digital attacks are categorized as fake in the final classification procedure, the significant differences between these attack types increase the intra-class distances. Current methods are tailored to address either physical or digital threats specifically. UAD algorithms and baselines for evaluating such algorithms are urgently needed.

Striving to propel advancements in the research community regarding UAD, we address the issues analyzed above through the following two aspects: (1) We collected and published a large-scale Unified Physical-Digital Attack dataset named UniAttackData~\cite{fang2024unified}. Compared to current unified datasets, it has several advantages, such as the complete attack types of each ID, the most advanced forgery methods, and the amount of data. (2) We establish a broader and more valuable testing protocol, which emphasizes evaluating the generalization ability of UAD algorithms. (3) Based on this dataset, we successfully held the \textbf{\textit{Unified Physical-Digital Attack Detection Challenge at CVPR2024}}, which attracted 136 teams worldwide. The top three teams achieved results significantly surpassing our baseline. A summary containing the names of the team and affiliations who reached the final phase is shown in Tab.~\ref{Tab1_ranks}. 

To sum up, the contributions of this paper are summarized as follows: 
\begin{itemize}
	\setlength{\itemsep}{1.0pt}
	\item
	We describe the design of the Unified Physical-Digital Attack Detection Challenge at CVPR2024.
	\item
	We organized this challenge around the UniAttackData, proving the suitability of such a resource for boosting research on the topic.
	\item
	We report and analyze the solutions developed by participants.
    \item
    We highlight critical factors in detecting both physical and digital by examining the top-ranked algorithms and suggest future research directions through this competition.
\end{itemize}

\section{Related Work}
\subsection{Face Anti-spoofing datasets for Challenges}

\textbf{Print-Attack}~\cite{Print-Attack_dataset} addresses the prevalent issue of bypassing 2D FR systems using spoofed photographs. It comprises 400 samples: 200 genuine accesses and 200 videos using printed photos across 50 identities. 
\textbf{Replay-Attack}~\cite{Replay-Attack_dataset} is a dedicated resource focusing on two-dimensional PAs, which includes 1,300 video clips featuring various attack types with 50 subjects, recorded under controlled and adverse lighting conditions. 
\textbf{OULU-NPU}~\cite{2Ddataset2_oulu} is a public PAD database designed to test the generalization of FR systems across different environments like lighting and background, various smartphones, and presentation attack instruments. It includes 5,940 high-resolution videos of 55 subjects, recorded in three environments with six smartphone models. 

\textbf{CASIA-SURF}~\cite{zhang2020casia} constitutes a large-scale resource that includes data from $1,000$ distinct subjects, featuring $21,000$ videos per subject, with each encompassing three modalities: RGB, Depth, and Infrared images. 
\textbf{CASIA-SURF CeFA}~\cite{liu2021casia} is a pioneering resource that explores ethnic bias in FAS by encompassing data from three ethnicities, three modalities, and 1,607 individuals, complete with 2D and 3D attack types. 

\textbf{CASIA-SURF HiFiMask}~\cite{liu2022contrastive} is a large-scale, high-fidelity mask dataset introduced to address the limitations of current 3D mask attack detection benchmarks. It contains over $54,600$ videos from 75 individuals wearing 225 realistically crafted masks using seven different sensor types. 
\textbf{CASIA-SURF SuHiFiMask}~\cite{fang2023surveillance} is a large-scale dataset designed for advancing FAS research. It specializes in protecting FR's security in distant surveillance scenarios and contains attack data from 101 individuals across various age ranges in 40 different surveillance settings. 

\textbf{UniAttackData}~\cite{fang2024unified} is the largest known dataset that unifies physical-digital attacks, involving each of 1,800 subjects with 2 physical and 12 digital attacks, which applied by SOTA attack methods over the last three years.

\subsection{Face Anti-spoof Challenges}
Since 2019, the Institute of Automation of the Chinese Academy of Sciences (CASIA) has held a series of FAS competitions based on the International Conference on Computer Vision. 

The first competition \textbf{Multi-modal Face Anti-spoofing Attack Detection Challenge at CVPR-2019}~\cite{liu2019multi} aimed to advance research in PAD by utilizing the large-scale dataset, CASIA-SURF. The challenge, held during CVPR2019, drew participation from over 300 global entities, with a majority from industry in the final stage, highlighting the practical significance of FAS in real-life applications. 

The second competition \textbf{Cross-ethnicity Face Anti-spoofing Recognition Challenge at CVPR-2020}~\cite{liu2021cross} addresses racial bias in FR systems, leveraging CASIA-SURF CeFA. The challenge featured separate tracks for single-modal (RGB) and multi-modal (combining with depth and infrared) FAS techniques. 340 teams participated in the development phase, with 11 and 8 teams advancing to the final evaluation for the single-modal and multi-modal tracks, respectively. The competition served as a catalyst for research focused on mitigating racial bias in facial spoofing detection.

The third competition \textbf{3D High-Fidelity Mask Face Presentation Attack Detection Challenge at ICCV-2021}~\cite{liu20213d} addressed the security risks associated with advanced 3D masks against FR systems. The challenge utilized the HiFiMask dataset featuring diverse and established a demanding testing protocol to measure algorithm performance, particularly distinguishing real vs. fake faces across varying conditions. The competition attracted 195 teams. Notably, many finalists hailed from industry backgrounds, underscoring the practical importance of this research for real-world applications.

The fourth competition \textbf{Surveillance Face Presentation Attack Detection Challenge at CVPR-2023}~\cite{fang2023surveillancechallenge} was conducted to enhance the security of FR systems in surveillance settings, particularly in long-distance scenarios by leveraging a large-scale SuHiFiMask dataset. A total of 180 teams participated, with 37 advancing to the finals. 

\section{Challenge Overview}
\label{sec:Overview}

\subsection{UniAttackData Dataset}
As far as we know, the UniAttackData~\cite{fang2024unified} is the largest unified physical-digital attack dataset, with a total of $28,706$ videos of $1,800$ subjects. It encompasses $1,800$ live face videos, $5,400$ videos showcasing PAs, and $21,506$ videos with DAs. Complete attack types for each ID are constructed to maintain consistency across physical and digital dimensions. This procedure avoids leading models inadvertently focusing on features unrelated to FAS tasks.

To facilitate the use of the dataset by the participating teams, the following pre-processing steps were carried out: (1) We clip and crop the facial part of the image from the original videos. (2) We choose one frame of each video and reconstruct the file's names, like\textit{ train/000001.png}, to hide the attack clues, like ethics or types.

\subsection{Challenge Protocol and Data Statistics}
To thoroughly evaluate the performance of UAD frameworks, we established two distinct protocols within the UniAttackData framework. According to Tab.~\ref{Tab2_protocol}, Protocol 1 is designed to scrutinize performance across unified attack tasks. Protocol 2, on the other hand, is tailored to assess algorithmic generalization across "unseen" attack types. Employing a "leave-one-type-out testing" strategy, we further divide Protocol 2 into two sub-protocols, which enables a thorough investigation into the generalization capacity of FAS mechanisms against a broad and evolving threat landscape.

\begin{table}[]
\scalebox{0.9}{
\begin{tabular}{|c|c|cccc|c|}
\hline
\multirow{2}{*}{Protocal} & \multirow{2}{*}{Class} & \multicolumn{4}{c|}{Types}                                                                   & \multirow{2}{*}{Total} \\ \cline{3-6}
                          &                        & \multicolumn{1}{c|}{Live} & \multicolumn{1}{c|}{Phys} & \multicolumn{1}{c|}{Adv}   & Digital &                        \\ \hline
\multirow{3}{*}{P1}       & train                  & \multicolumn{1}{c|}{3000} & \multicolumn{1}{c|}{1800} & \multicolumn{1}{c|}{1800}  & 1800    & 8400                   \\ \cline{2-7} 
                          & eval                   & \multicolumn{1}{c|}{1500} & \multicolumn{1}{c|}{900}  & \multicolumn{1}{c|}{1800}  & 1800    & 6000                   \\ \cline{2-7} 
                          & test                   & \multicolumn{1}{c|}{4500} & \multicolumn{1}{c|}{2700} & \multicolumn{1}{c|}{7106}  & 7200    & 21506                  \\ \hline
\multirow{3}{*}{P2.1}     & train                  & \multicolumn{1}{c|}{3000} & \multicolumn{1}{c|}{0}    & \multicolumn{1}{c|}{9000}  & 9000    & 21000                  \\ \cline{2-7} 
                          & eval                   & \multicolumn{1}{c|}{1500} & \multicolumn{1}{c|}{0}    & \multicolumn{1}{c|}{1706}  & 1800    & 5006                   \\ \cline{2-7} 
                          & test                   & \multicolumn{1}{c|}{4500} & \multicolumn{1}{c|}{5400} & \multicolumn{1}{c|}{0}     & 0       & 9900                   \\ \hline
\multirow{3}{*}{P2.2}     & train                  & \multicolumn{1}{c|}{3000} & \multicolumn{1}{c|}{2700} & \multicolumn{1}{c|}{0}     & 0       & 5700                   \\ \cline{2-7} 
                          & eval                   & \multicolumn{1}{c|}{1500} & \multicolumn{1}{c|}{2700} & \multicolumn{1}{c|}{0}     & 0       & 4200                   \\ \cline{2-7} 
                          & test                   & \multicolumn{1}{c|}{4500} & \multicolumn{1}{c|}{0}    & \multicolumn{1}{c|}{10706} & 10800   & 26006                  \\ \hline
\end{tabular}
}
\caption{Two protocols used in this challenge.}
\label{Tab2_protocol}
\end{table}

\subsection{Challenge Process and Timeline}
The challenge was held on the CodaLab platform, including two phases as follows:

\textbf{Development Phase:}
(Start: Feb. 1st, 2024 – Ended: Feb. 22nd). During this phase, participants can train their models with labeled training data and predict scores on unlabeled development data. Two data subsets include the same type of attacks (unified attacks for phase 1, digital for phase 2.1, and physical for phase 2.2). Participants can submit their predictions on the development set and get timely feedback via leader board.

\textbf{Final Phase:}
(Start: Feb. 23rd – Ended: Mar.3rd). In the final phase, we released the development set labels for participants to refine their models for the test set, while also making the unlabeled test data available. Teams are required to predict outcomes for the test data and submit these to CodaLab. It is crucial that models are trained solely on the training set, without access to the development or test set data. Note that on the CodaLab platform, the last submission is considered the final entry.

The final ranking of participants was obtained from the submissions' performance in the testing sets. To be eligible for prizes, winners had to publicly release their code under a license and provide a fact sheet describing their solution.

\subsection{Evaluation Metrics}
In evaluating the performance for this challenge, we adopted the ISO/IEC 30107-3 metrics, which include the Attack Presentation Classification Error Rate (APCER), the Normal/Bona Presentation Classification Error Rate (NPCER/BPCER), and the Average Classification Error Rate (ACER). These metrics quantify the detection system's accuracy in identifying live and fake face presentations. The determination of ACER for the test set relies on the Equal Error Rate (EER) established during the development phase. Additionally, we used the Area Under Curve (AUC) metric as an additional performance measure, assessing model discrimination between fake and real samples across thresholds. Rankings were primarily based on the Average Classification Error Rate (ACER), with AUC as a secondary criterion, to thoroughly evaluate each algorithm’s efficacy against spoofing attacks.

\section{Description of solutions}

\subsection{MTFace} 

\begin{figure}[h!]
    \centering
    \includegraphics[width=1.0\linewidth]{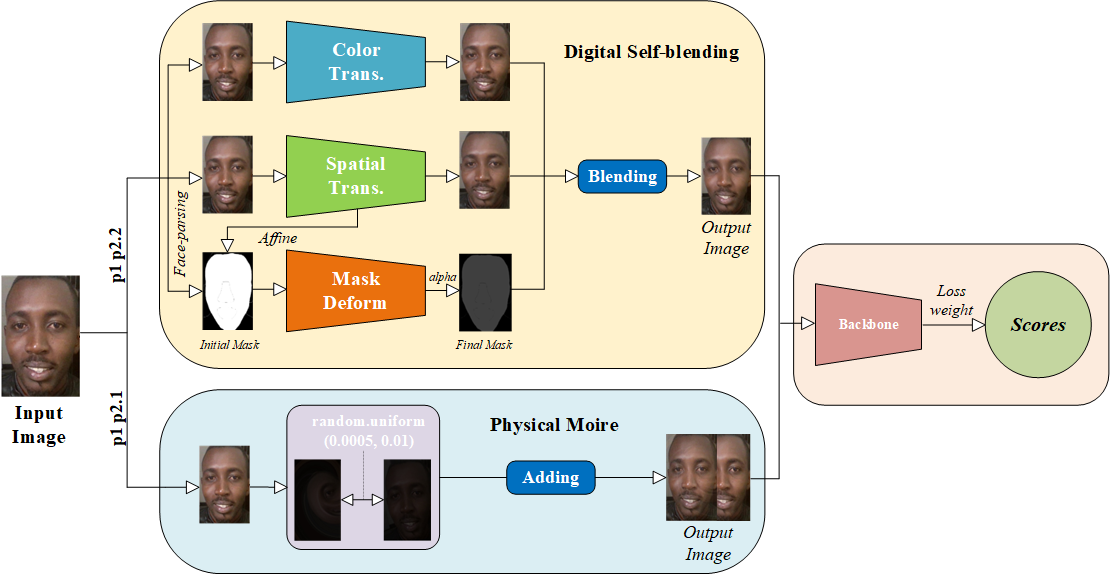}
    \caption{The overall pipeline of MTFace's method. The moiré augmentation improves the model’s PAD ability, while self-blending augmentation improves the model’s DAD ability.}
    \label{Fig3_1_MTFace}
\end{figure}

Due to the significant disparity between samples applying physical attacks and those applying digital attacks, team MTFace proposed a framework named Optimized Data Augmentation for Comprehensive Face Attack Detection Across Physical and Digital Domains. This framework focuses prominently on data augmentations and balanced training loss. The overall pipeline, including data augmentation and loss-balancing training, is shown in Fig.~\ref{Fig3_1_MTFace}.

During pre-processing, all uncropped images are face-detected and cropped 20 pixels outward by the bound box. Finally, all live data in the training set is parsed to obtain the face mask area for further use in subsequent self-blending methods. As shown in Fig.~\ref{Fig3_1_MTFace}, a moiré data augmentation and a self-blending augmentation are applied to the different subsets based on the diverse structure of protocols. MTFace analyzes that screen replay will cause moiré patterns on the surface texture of the image. Thus, the moiré data augmentation method is designed and applied to protocol p1 and p2.1 so that the characteristics of physical attacks are introduced, improving the model’s ability to detect physical attacks across domains. Second, inspired by the work~\cite{1MTFace2_selfblend}, a self-blending augmentation method is defined. For protocols p1 and p2.2, extra live-based data has been added with digital attack features by color, spatial transformation, and mask deformation. To achieve a more balanced training approach in response to the varying ratios of bona and spoof samples across three protocols, MTFace adjusted the cross-entropy loss weights accordingly. For protocol p1, they maintained an equal loss weight ratio of live to fake samples at 1:1, aiming for an even-handed learning process. In protocol p2.1, they adopted a 5:1 loss weight ratio, favoring live data to encourage the model to learn live features more effectively. For protocol p2.2, they applied a 2:1 loss weight ratio to better equilibrate the impact of different sample types on the model's learning. This led to a marked enhancement in their experimental outcomes. Finally, ResNet-50 is set as the backbone, and the ImageNet pre-train weights are loaded. Their promising experiment results have beaten all other competitors in this challenge.

\subsection{SeaRecluse} 

The team SeaRecluse presented a solution titled "Cross-domain Face Anti-spoofing in Unified Physical-Digital Attack Dataset." The team's approach included several key steps and techniques: SCRFD is first applied to uncropped images in the training set for face cropping. For three protocols, the dataset partitioning methods differ: $80\%$ of the p1, $60\%$ of p2.1, and $40\%$ of p2.2's training data are used along with the validation data for training, while the remaining data serves as the validation set. Regarding data augmentation and preprocessing, cropped images undergo no additional processing, whereas uncropped images are supplemented with data through loose and tight facial cropping. For each of the three tasks, distinct data augmentation measures are taken based on their respective focuses: Task P1 does not perform any enhancement operations; for P2.1, to balance the ratio of real face samples, real face data is downsampled and edge blank pixels are filled to reach a fixed size, effectively tripling the amount of real face data; in contrast, for P2.2, similar augmentation strategies are adopted for fake face data, introducing 4x and 8x downsampling, thereby increasing the fake face data volume by sevenfold. During the data review stage, some images' incorrect aspect ratios were rectified and restored to their original proportions. All tasks implement normal data augmentation operations, with Task P2.1 also incorporating Gaussian blur. SeaRecluse chose ConvNeXt V2 backbone considering the competition requirements and local training resources for training. During the training phase, Image CutMix and label smoothing~\cite{liu2022convnet} techniques were leveraged to enhance the model's generalization capability.

\subsection{duileduile} 

\begin{figure}[t!]
    \centering
    \includegraphics[width=1.0\linewidth]{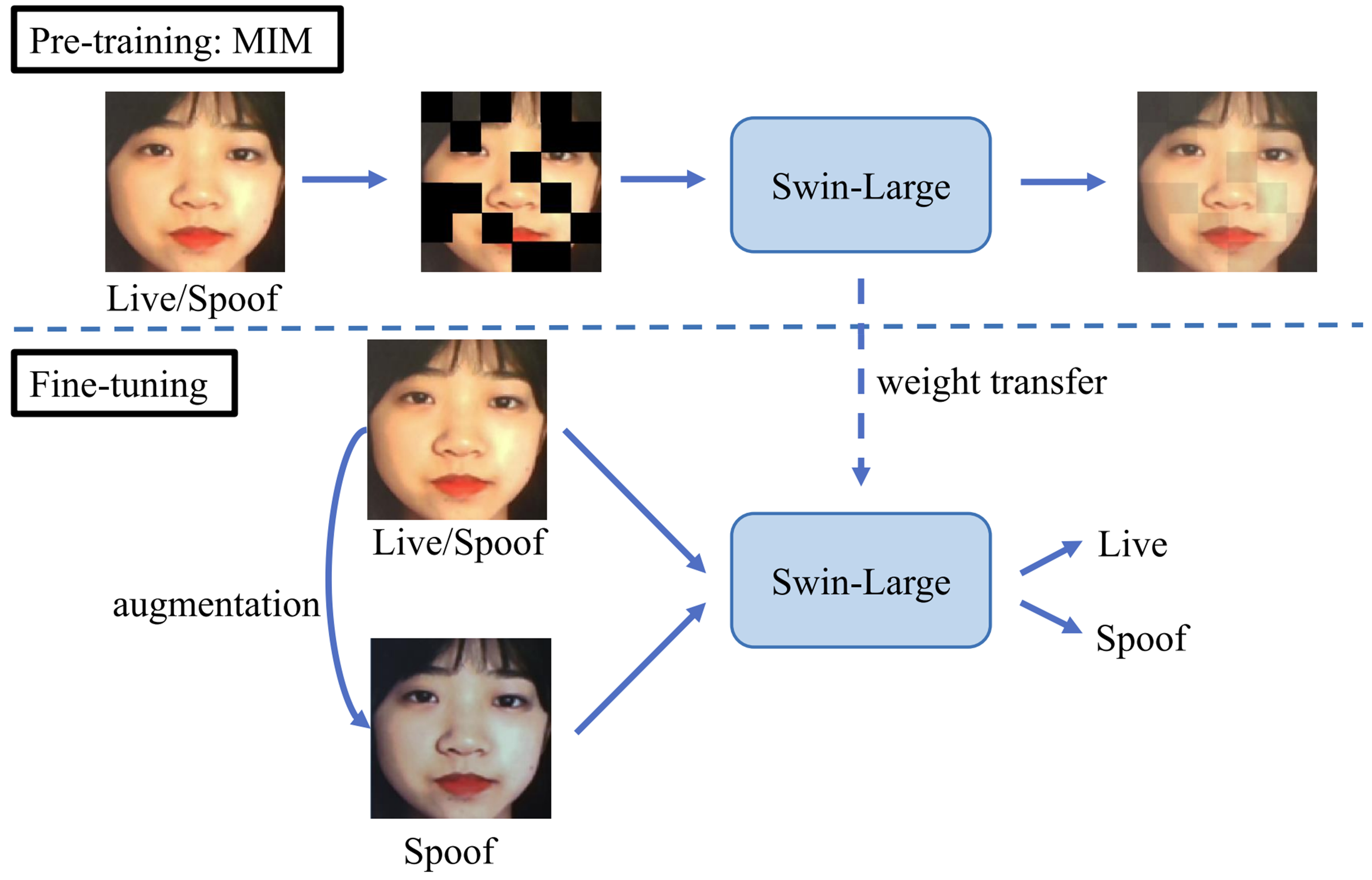}
    \caption{The pipeline of Team duileduile's framework with pre-training stage and fine-tuning stage.}
    \label{Fig3_3_duileduile}
\end{figure}

The team duileduile has devised a two-stage method integrating pre-training and fine-tuning techniques, employing the Swin Transformer model~\cite{3Insig1_swim} as its backbone. This approach showcases a robust framework for generalizing the detection of unified face representations, and the solution pipeline is shown in Fig.~\ref{Fig3_3_duileduile}.

The backbone model, the Swin-Large model~\cite{3Insig1_swim}, extracts features resulting in 1536-dimensional vectors, which, coupled with the image's dimensional attributes, aid in differentiating between live and spoof images. Both pre-training and fine-tuning stages follow identical configurations across all protocols. This self-contained strategy yields a competitive edge in cross-attack-type scenarios and simplifies the transplanting of the method across various baselines. The simMIM strategy~\cite{3Insig2_simMIM} is utilized with unlabeled training and development sets at the pre-training stage, involving segmenting an image into non-overlapping patches, concealing portions and prompting the model to infer the complete image. This masked image modeling (MIM) methodology fosters the model's proficiency in feature extraction from incomplete visual information, which is helpful for protocol 2 in the testing phase. During the fine-tuning stage, they introduce a tailored data augmentation sequence. The Gaussian Noise augmentation represents digital threats, while ColorJitter and screen display simulations, including moiré pattern overlays and gamma correction, are used to mimic physical attack types. These sequential augmentations are applied exclusively to training samples with designated probabilities.

\subsection{BSP-Idiap} 

\begin{figure}[t!]
    \centering
    \includegraphics[width=0.99\linewidth]{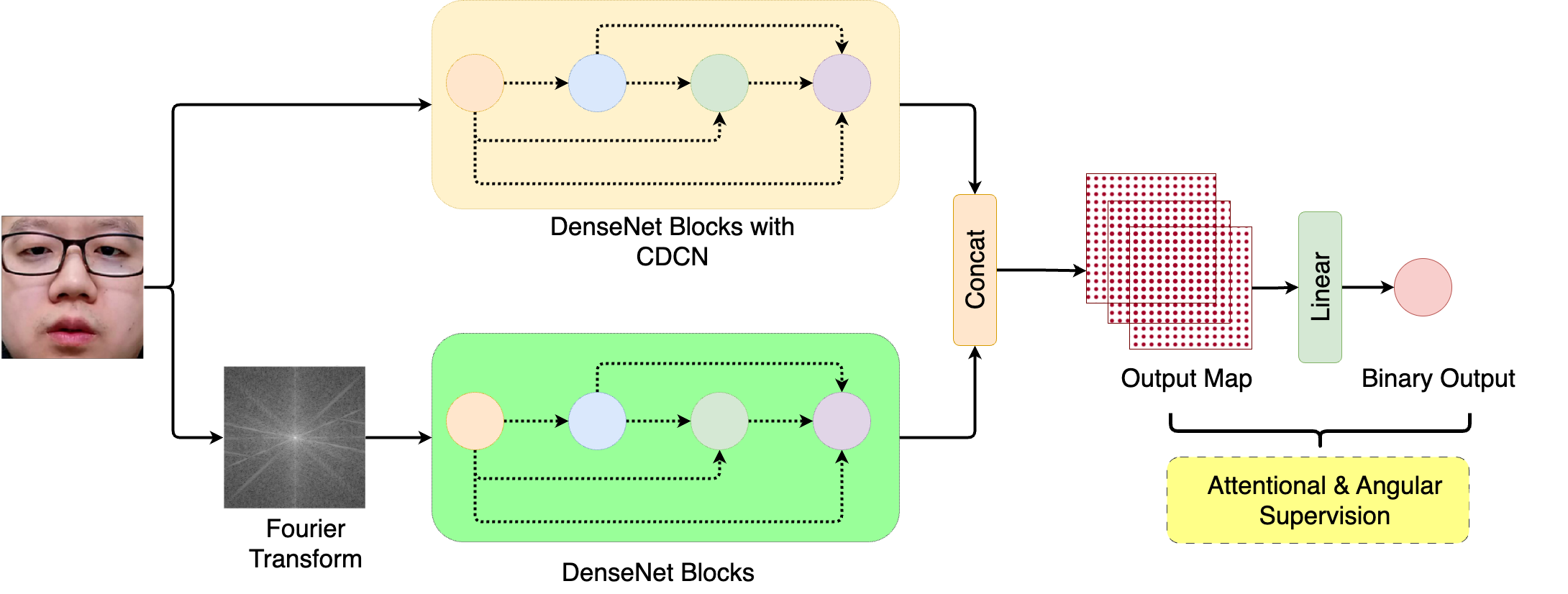}
    \caption{This figure shows the DBPixBiS's architecture, the feature map, and the final outputs are supervised by angular loss.}  \label{Fig4_4_BSP_Idiap}
\end{figure}

The Dual Branch Pixel-wise Binary Supervision (DBPixBiS) algorithm, presented by the BSP-Idiap team, is a novel FAS solution that extends their previous work DeepPixBiS~\cite{george2019deep}. As shown in Fig.~\ref{Fig4_4_BSP_Idiap}, their DBPixBiS utilizes a dual-branch network architecture that takes Fourier transform representations as input to capture both spatial and frequency domain properties of images. In the RGB branch of the model, central difference convolutional blocks~\cite{yu2020searching} are employed to process the visual information differently from standard convolutional operations. A vital aspect of the method lies in its pixel-wise binary supervision, where each pixel in the feature maps is guided by an attentional angular margin loss~\cite{hossain2020deeppixbis} function during the training phase. This supervision strategy regularizes the learning process, reducing the over-fitting problem and promoting better generalization across diverse spoofing scenarios. The architecture includes an additional branch tailored for Fourier-transformed inputs to detect spoofing artifacts in the frequency domain. During inference, the mean value of the resulting feature map is used as the final spoofing detection score.

\subsection{VAI-Face} 

\begin{figure}[!ht]
    \centering
    \includegraphics[width=1.0\linewidth]{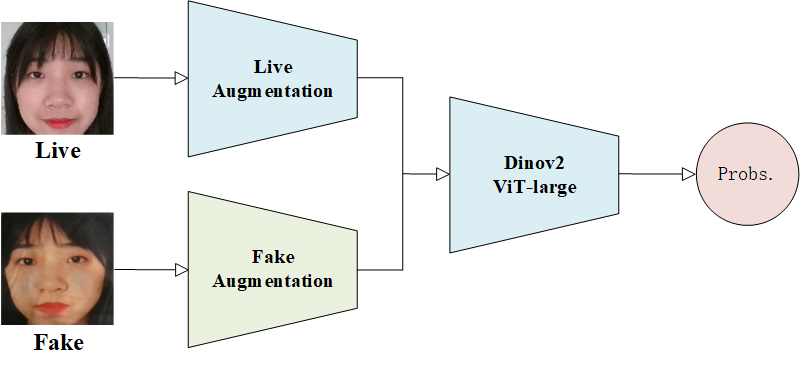}
    \caption{The overall structure of team VAI-Face's solution.}
    \label{Fig5_5_VAI_Face}
\end{figure}

The team VAI-Fac's approach is distilled into Fig.~\ref{Fig5_5_VAI_Face}, briefly outlining their methodology. For this task, they utilized the Dinov2 Vision Transformer (ViT) large model~\cite{5VAIFace_dinov2}. A key element of their strategy is the application of disparate augmentation techniques to live and fake images to enhance model performance. Live images are treated with RandomResizedCrop and HorizontalFlip, while fake images are subjected to a more extensive augmentation suite that includes distortions, blurs, and custom cutouts to simulate the anomalies often found in spoofed images. Their learning strategies include utilizing OneCycleLR with finely-tuned hyperparameters and label smoothing, emphasizing precision in enhancing the model's learning efficiency. The VAI-Face team further bolstered their model's robustness with mixup augmentation and opted for the ADAN optimizer. In summary, the VAI-Face team's methodology showcases a considered balance between sophisticated data augmentation strategies, a thorough hyperparameter tuning process, and leveraging the capabilities of a powerful ViT model, making it a standout contribution in the arena of UAD.

\subsection{L\&L\&W} 

The team L\&L\&W has introduced a method centered on patch-based feature learning and incorporating novel attention and sampling techniques for FAS. L\&L\&W's methodology starts with extracting small patches from images, which are then processed to learn discriminative features. They employ Centre Difference Attention to capture fine-grained intrinsic features, highlighting the subtle cues essential for spoof detection. Furthermore, they have developed a High-Frequency Wavelet Sampler, focusing on high-frequency components to robustly identify forgery traces in the spatial domain. This indicates a dual focus on spatial and frequency domains for a comprehensive feature extraction strategy. In validation and testing, they create 36 crops of an image and use the average of the predictions as the final score. This approach suggests an emphasis on capturing varied aspects of the images to enhance the prediction reliability. 

\subsection{SARM} 

Through analysis, the SARM team employed a multi-attention (MAT) network structure to train detection models at each stage, noticing the variation of protocols. Their approach consists of two stages: initially, training pre-trained detection models for each phase using supervised contrastive learning; subsequently, fine-tuning these pre-trained models in the second stage to generate FAS models. During training, they combine cross-entropy loss and supervised contrastive learning loss. While employing some general training strategies and the Adam optimizer for the relatively straightforward protocol P1, they adopted a label-flip augmentation strategy to generate more synthetic training data for P2.1 and P2.2. Specifically, they utilize image style transformation techniques based on the OpenCV library to convert real face images into synthetic face images, labeling these generated synthetic images as fake. This strategy significantly improves performance and reduces the domain gap between the training and testing sets.

\subsection{M2-Purdue} 

\begin{figure}[!ht]
    \centering
    \includegraphics[width=1.0\linewidth]{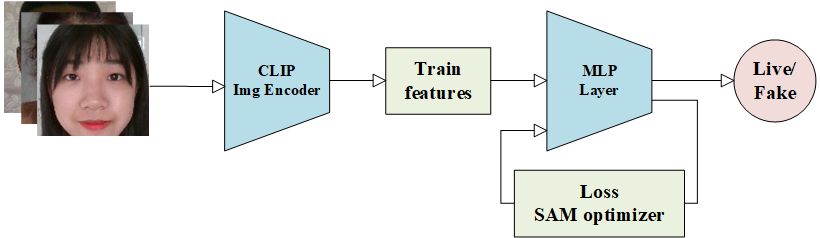}
    \caption{The overall framework of team M2-Purdue's algorithm.}
    \label{Fig8_8_Purdue}
\end{figure}

To enhance the study in UAD, the M2-Purdue team proposed a method that blends feature extraction using CLIP with a Multi-Layer Perceptron (MLP) based classification system, namely Robust Face Attack Detection. The framework of their approach is depicted in Fig.~\ref{Fig8_8_Purdue}.

Their methodology commences with pre-processing, which resizes all images to 224x224 pixels to standardize the input data. Following this, the team utilizes the CLIP~\cite{8Purdue1_clip} model's image encoder to extract detailed feature representations of the visual content. Next, the MLP, composed of three layers, serves as the pivot of their classification mechanism. This MLP is fine-tuned using a composite loss function that combines the Conditional Value at Risk (CVAR) loss and the Area Under the ROC Curve (AUC) loss, weighted by a parameter  $\lambda$. This fusion is described as Eq.~\ref{Eq1_3_8_loss}: 
\begin{equation}
  \operatorname{\mathcal{L}}=\lambda \mathcal{L}_{\mathrm{CVAR}}+(1-\lambda) \mathcal{L}_{\mathrm{AUC}}
  \label{Eq1_3_8_loss}
\end{equation}
where $\lambda$ balances the focus between the two losses, harnessing their strengths. Also, the Adaptive Moment Estimation (Adam) optimizer handles the model's optimization, further refining the training process.

\subsection{Cloud Recesses} 

\begin{figure}[!ht]
    \centering
    \includegraphics[width=1.0\linewidth]{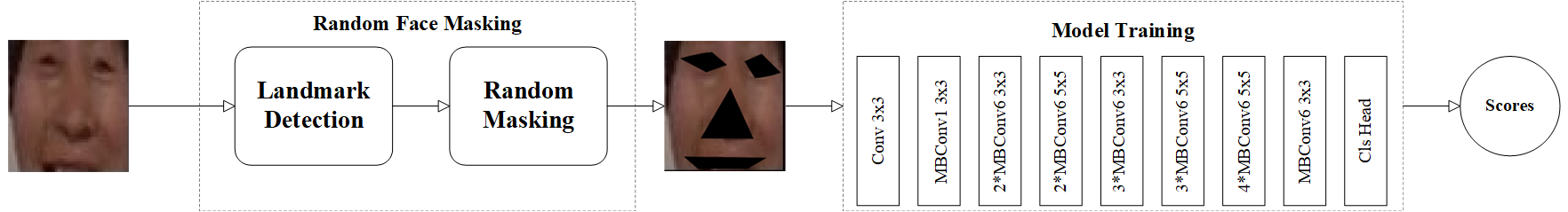}
    \caption{The overview of team Cloud Recesses's scheme. Face landmarks are first detected and used to guide random face masking. Then, the EfficientNet model is trained for classification.}
    \label{Fig9_9_CloudReccesses}
\end{figure}

The Cloud Recesses team has advanced a technique titled "Random Masking for Face Anti-Spoofing Detection." They proposed an innovative scheme, depicted in Fig.~\ref{Fig9_9_CloudReccesses}.
The central premise of their method pivots on data augmentation through the deliberate obscuration of critical facial features. Initially, face landmarks are detected to inform the random masking process, intentionally veiling key areas such as the eyes, nose, and mouth. Subsequently, these modified images are channeled into the EfficientNet architecture for model training, targeting the enhancement of the model's predictive acumen on validation sets. RetinaFace is employed for face detection in data pre-processing, yielding a 256x256 cropped face image. After that, dlib is utilized to detect 68 facial landmarks, outlining essential face components. During training, random occlusion of 3-5 regions per image is executed, attenuating the impact of identifiable facial features on the training outcome.

\subsection{ImageLab} 

\begin{figure}[t!]
    \centering
    \includegraphics[width=1.0\linewidth]{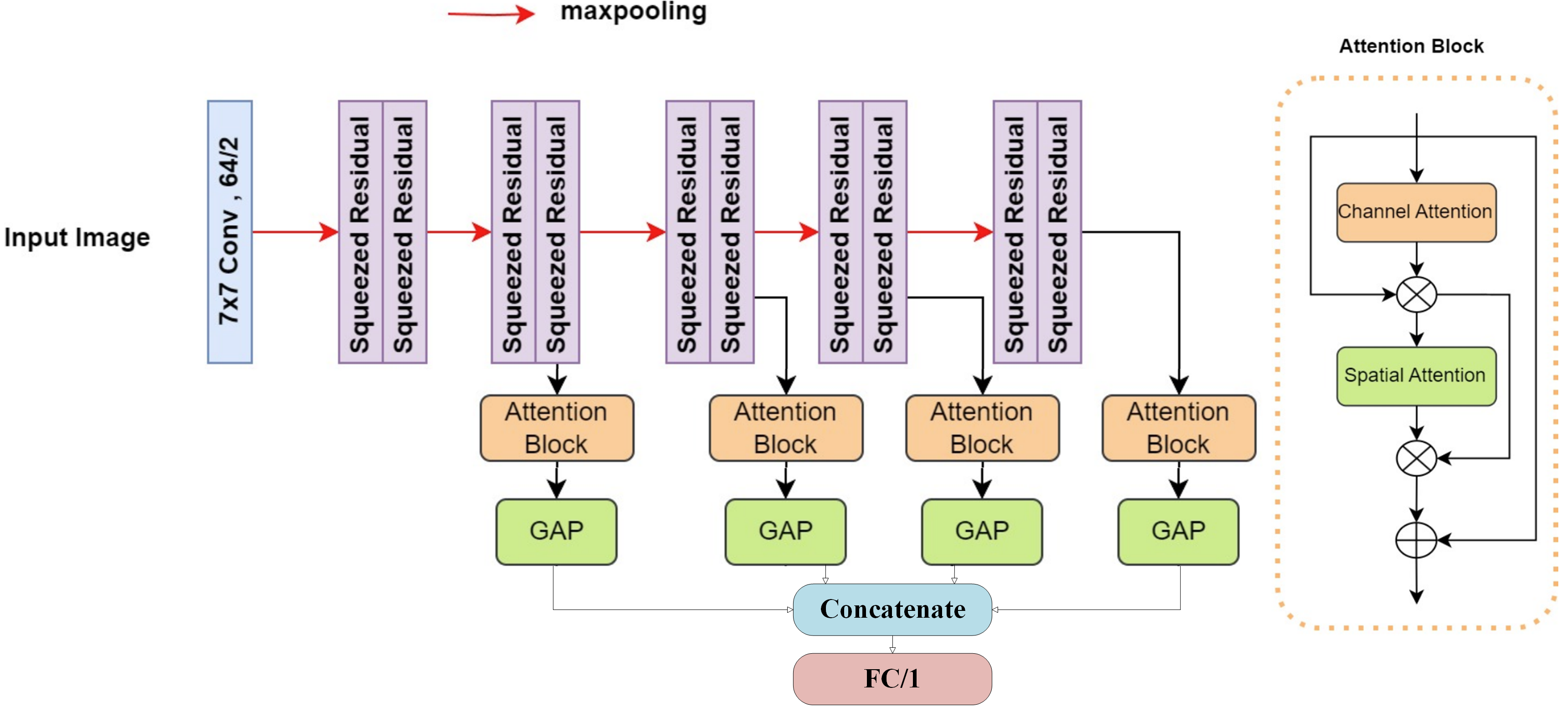}
    \caption{The overall architecture of Multiattention-Net proposed by team ImageLab.}
    \label{Fig10_10_ImageLab}
\end{figure}

The Image Lab team unveiled a sophisticated framework named Multiattention-Net for face anti-spoofing. Their proposed solution is elucidated in Fig.~\ref{Fig10_10_ImageLab}.
The Multiattention-Net initiates with a 7x7 convolutional layer to grasp complex local patterns within the input image. The network is then deepened with ten modified squeezed residual blocks that systematically downsample the information through max pooling to extract more abstract and global features. Spatial information extraction is performed at various levels of these blocks, followed by applying a dual attention block~\cite{10ImageLab4_dualattention} that accentuates critical features within the spatial domain. A global average pooling (GAP) strategy is employed to diminish feature dimensions, and the outputs from different spatial levels are concatenated before being passed on to a fully connected layer. The training regimen is bolstered by a Binary Focal Cross entropy loss function, delicately weighted with a class balancing factor $\alpha$ and a focusing parameter $\lambda$, designed to tackle the challenges posed by imbalanced datasets by imposing more significant penalties on incorrect predictions, particularly for the minority class. The equation can be represented as Eq.~\ref{Eq2_3_10_loss}:
\begin{equation}
  \mathcal{L}(y, \hat{y})=-\alpha \cdot(1-\hat{y})^{\gamma} \cdot \log (\hat{y})-(1-\alpha) \cdot \hat{y}^{\gamma} \cdot \log (1-\hat{y})
  \label{Eq2_3_10_loss}
\end{equation}
where $\mathcal{L}$ is the binary focal cross-entropy loss, $y$ is the ground truth label (0 or 1), $\hat{y}$ is the predicted probability, $\alpha$ is the class balancing factor (0.25 if apply class balancing=True), $\gamma$ is the focusing parameter (3 in this case)

\subsection{BOVIFOCR-UFPR} 

\begin{figure}[!ht]
    \centering
    \includegraphics[width=1.0\linewidth]{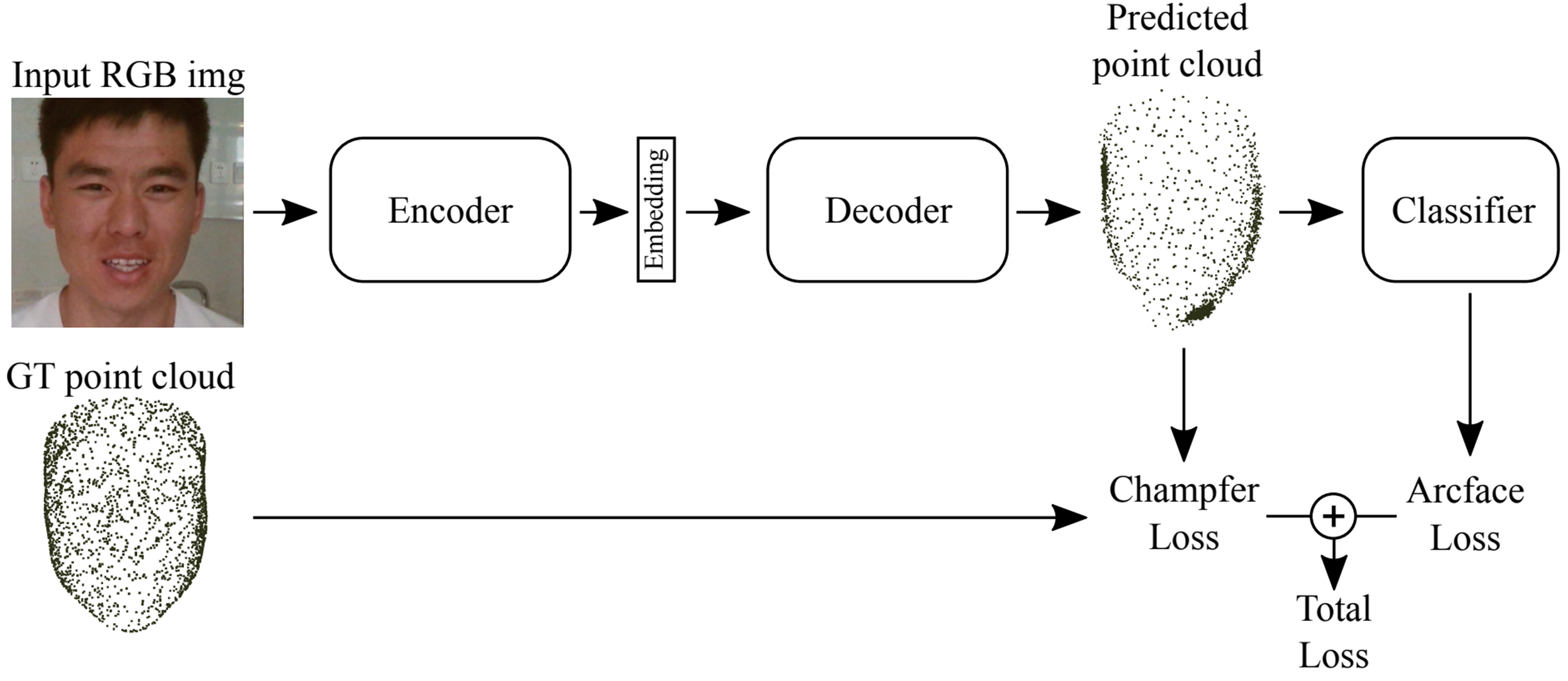}
    \caption{General architecture of BOVIFOCR-UFPR's method. The model receives pairs of RGB face images and its corresponding ground truth 3D point cloud. The predicted point cloud is also fed to a final classifier.}
    \label{Fig11_11_BOVI}
\end{figure}

The BOVIFOCR-UFPR team crafted a sophisticated approach to UAD by leveraging 3D face reconstruction and angular margin loss, as portrayed in their method overview in Fig.~\ref{Fig11_11_BOVI}. 
The proposed architecture draws inspiration from 3DPC-Net~\cite{PAD6_3dpcnet} and pivots on an encoder-decoder setup. First, faces are detected, aligned, and cropped uniformly in the preparatory stages. The ground truth 3D point clouds corresponding to these images are initially reconstructed using a high-quality reconstruction method~\cite{11BOVI3_hrn3d}. Then, the encoder, built upon a ResNet-50 backbone, extracts high-level features from the input RGB images. The decoder transforms these features into a 3D point cloud representing the facial structure, subsequently used to distinguish between live and spoof faces. During the training phase, the Arcface loss function is used to refine its discriminative power, and the fine-tuned Stochastic Gradient Descent optimizer collectively orchestrates the learning process. Furthermore, the Chamfer loss function, known for its $O(N^2)$ complexity, is employed alongside the Arcface loss function, which bears a complexity of $O(CD)$, with $C$ denoting the number of classes and $D$ the dimensionality of the encoder's feature output.

\subsection{Inria-CENATAV-Tec} 

\begin{figure}[!ht]
    \centering
    \includegraphics[width=1.0\linewidth]{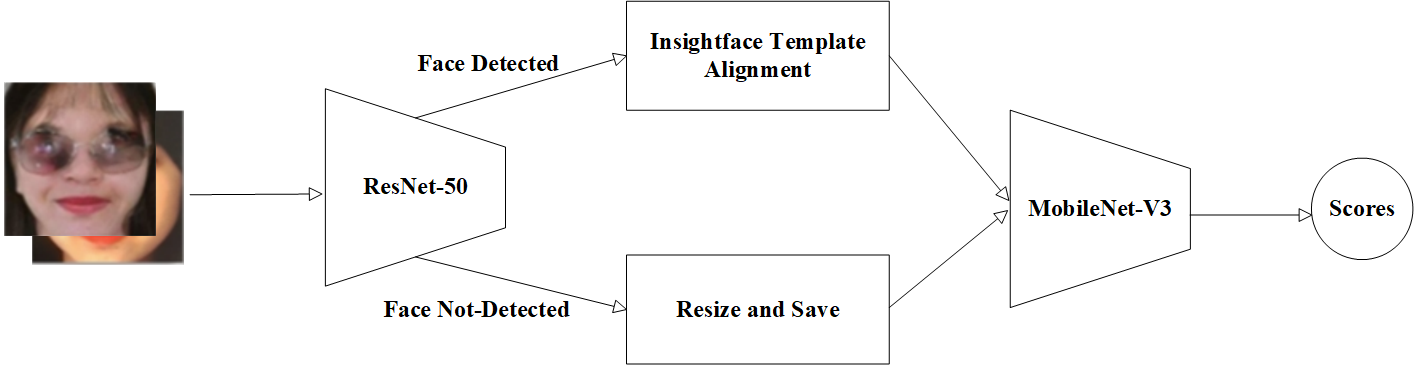}
    \caption{The approach of team Inira-CENATAV-Tec's Pre-processing and UAD stages.}
    \label{Fig12_12_inria}
\end{figure}

The team Inria-CENATAV-Tec refined UAD with their contribution, "MobileNetV3-spoof with hyperparameter tuning." Their methodology is encapsulated in Fig.~\ref{Fig12_12_inria}, which details a flowchart of their pre-processing and UAD approach. In the pre-processing phase, the team employs ResNet-50 for landmark detection to ascertain face positioning. If the face is detected, it is aligned using the insightface template; otherwise, the image is resized and saved in its original state. The team utilizes the MobileNetV3-large-1.25 backbone for feature extraction and spoof detection, aiming to balance model complexity with performance. The model training leverages a Stochastic Gradient Descent optimizer with a multi-step learning rate strategy. Additionally, various augmentation techniques are applied to the data before normalization, aligning with each protocol's mean and standard deviation. Their work aligns with the contemporary need for lightweight models that do not compromise the precision of FAS measures.

\subsection{Vicognit} 

\begin{figure}[!h]
    \centering
    \includegraphics[width=1.0\linewidth]{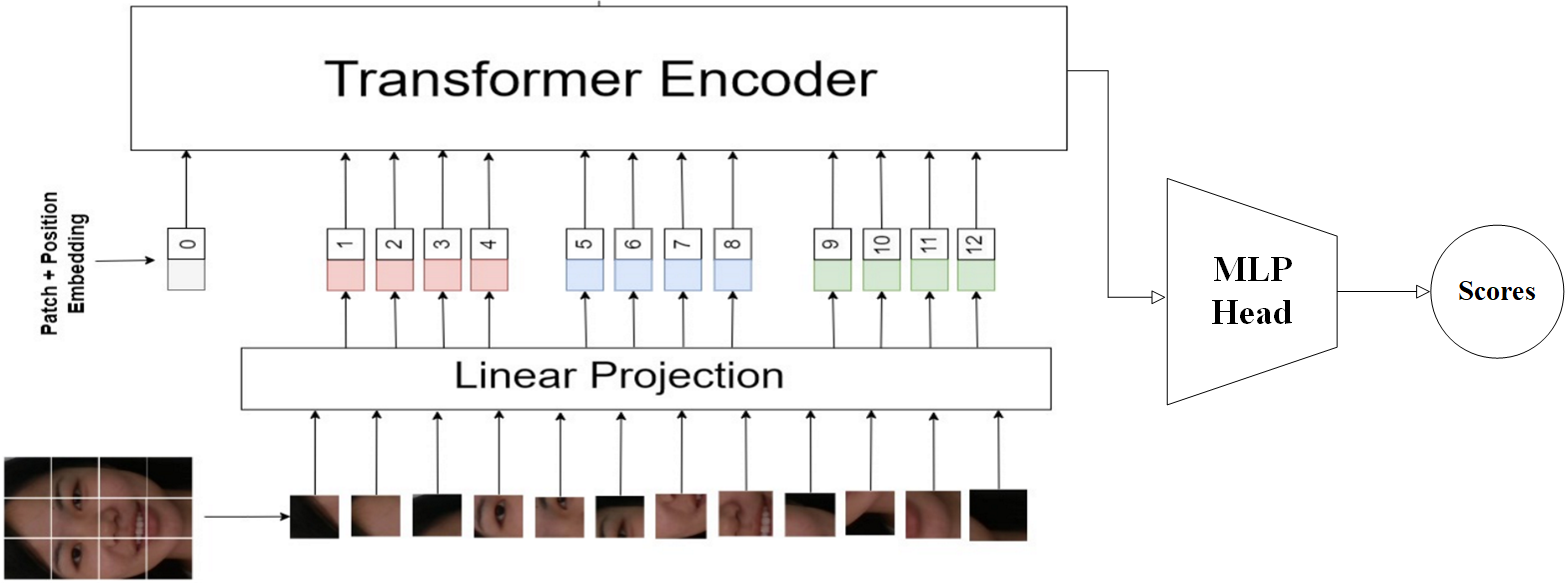}
    \caption{The approach of team Inira-CENATAV-Tec's Pre-processing and UAD stages.}
    \label{Fig13_13_vicognit}
\end{figure}

The team Vicognit presents "FASTormer: Leveraging Vision Transformers for Face Anti-Spoofing," a novel approach utilizing the Vision Transformer architecture for FAS. Their strategy is visualized in Fig.~\ref{Fig13_13_vicognit}, which outlines the architecture of the proposed method. During the training phase, the team meticulously tunes the hyperparameters, such as learning rate and weight decay, to ensure that the model's parameters are optimized effectively, leading to robust convergence and generalization. The Vicognit team's solution hinges on a transformer-based model, indicating their focus on harnessing transformers' capabilities to capture the intricate patterns necessary for distinguishing between bona and spoof face representations without reducing the dimensionality of the input data.

\section{Challenge Results}
\subsection{Challenge Results Report}
We adopted four metrics to evaluate the performance of the solutions: APCER, NPCER, ACER, and AUC. Please note that although we report performance for various evaluation measures, the leading metric is ACER. See Tab.~\ref{Tab3_results}, which lists the results and ranking of the top 13 teams; we can draw three conclusions: (1) The ACER performance of the top 3 teams was significantly higher than the other teams. (2) The first-rank team achieved the best results in ACER, AUC, and BPCER, while the fifth-rank team got the best APCER result. (3) The top 5 teams are from the industry, which indicates that developing UAD algorithms in real-world applications is crucial. (4) There is a considerable variation among teams concerning the ACER scores, inferring the desperation of the in-depth research on UAD in the Face Anti-Spoofing area.

\begin{table}[!h]
\scalebox{0.67}{
\begin{tabular}{|c|c|c|c|c|c|}
\hline
Rank & Team              & ACER(\%) & ACPER(\%) & BPCER(\%) & AUC(\%) \\ \hline
1    & MTFace            & \textbf{2.3396}   & 0.9259    & \textbf{3.7533}    & \textbf{99.6923} \\ \hline
2    & SeaRecluse        & 3.4369   & 0.3999    & 6.4737    & 96.5631 \\ \hline
3    & duileduile        & 5.5111   & 5.5185    & 5.5037    & 98.6830 \\ \hline
4    & BSP-Idiap         & 16.2263  & 9.3630    & 23.0698   & 96.3351 \\ \hline
5    & VAI-Face          & 17.1324  & \textbf{0.2593}    & 34.0055   & 87.7566 \\ \hline
6    & L\&L\&W           & 23.6949  & 11.8889   & 35.5009   & 81.0384 \\ \hline
7    & SARM              & 27.1958  & 0.5037    & 53.8879   & 98.3233 \\ \hline
8    & M2-Purdue         & 33.4651  & 0.8593    & 66.0709   & 86.4003 \\ \hline
9    & Cloud Reccesses   & 34.5701  & 1.8741    & 67.2661   & 71.6810 \\ \hline
10   & ImageLab          & 34.9434  & 5.6148    & 64.2717   & 76.6426 \\ \hline
11   & BOVIFOCR-UFPR     & 35.3620  & 6.8444    & 63.8795   & 73.7304 \\ \hline
12   & Inria-CENATAV-Tec & 37.3436  & 7.5259    & 67.1612   & 74.2011 \\ \hline
13   & Vicognit          & 52.1031  & 80.0000   & 24.2062   & 46.7025 \\ \hline
\end{tabular}
}
\caption{Team and results are listed in the final ranking of this challenge.}
\label{Tab3_results}
\end{table}

\subsection{Competition Summary and Future Work}

Through the challenge, we summarize the practical ideas for UAD: (1) For backbone networks, larger models with more parameters often perform better in complex real-world applications. (2) At the data level, pre-processing methods are crucial to enriching data variety, which helps avert model overfitting, while category balance enhances algorithmic stability. (3) Learning discriminative features from incomplete faces would be an effective path to UAD. In the following work, we will further improve the performance in the following aspects: (1) We will explore more efficient approaches by applying recent VLMs like CLIP~\cite{8Purdue1_clip} to guide the training process of UAD. (2) We will continue to create a more complete UAD dataset with more complete attack types and high-quality images. (3) We will try to develop a more advanced and authorized protocol for UAD tasks.

\section{Conclusion}
We organized the \textbf{\textit{Unified Physical-Digital Face Attack Detection Challenge at CVPR2024}} based on the UniAttackData dataset and running on the CodaLab platform. 133 teams registered for the competition, and 13 made it to the final stage and submitted their codes. In the final stage of the competition, the codes are verified and re-produced by the organizers, and the reproduced results are used for the final rankings. We first present the associated dataset, protocols, and evaluation metrics. Then, we review the solutions of the participating ranked teams and report the results of the final phase. Finally, we summarize the conclusions related to the challenges and point out effective methods for Unified Attack Detection by this challenge.
\vspace{1mm}
\section*{Acknowledgments}
This work was supported by the National Key Research and Development Program of China under Grant 2021YFE0205700, Beijing Natural Science Foundation JQ23016, the Science and Technology Development Fund of Macau Project 0123/2022/A3, and 0070/2020/AMJ, the Chinese National Natural Science Foundation Project 62276254, U23B2054, and the InnoHK program.
{
    \small
    \bibliographystyle{ieeenat_fullname}
    \bibliography{main}
}


\end{document}